\documentclass[letterpaper, 10 pt, conference]{ieeeconf}  %

\IEEEoverridecommandlockouts                              %

\usepackage{graphics} %
\usepackage{graphicx}
\usepackage{epsfig} %
\usepackage{mathptmx} %
\usepackage{times} %
\usepackage{amsmath} %
\usepackage{amssymb}  %
\usepackage{hyperref}
\usepackage{booktabs}
\usepackage{xcolor}
\usepackage{multirow}
\usepackage{array}
\usepackage{url}
\usepackage{cite}

\title{\LARGE \bf
TANGO: \underline{T}raversability-\underline{A}ware \underline{N}avigation with Local Metric Control for Topological \underline{Go}als
}

\author{Stefan Podgorski$^{*1}$, Sourav Garg$^{*1}$, Mehdi Hosseinzadeh$^{1}$, Lachlan Mares$^{1}$, Feras Dayoub$^{1}$, Ian Reid$^{1,2}$%
\thanks{$^{1}$Australian Institute for Machine Learning (AIML), The University of Adelaide, Australia.}%
\thanks{$^{2}$ Mohamed Bin Zayed University of Artificial Intelligence, UAE.}%
\thanks{*Equal contirbution.}%
}
\newcommand{\ourM}{TANGO}

\begin{document}

\maketitle
\thispagestyle{empty}
\pagestyle{empty}

\begin{abstract}
Visual navigation in robotics traditionally relies on globally-consistent 3D maps or learned controllers, which can be computationally expensive and difficult to generalize across diverse environments. In this work, we present a novel RGB-only, object-level topometric navigation pipeline that enables zero-shot, long-horizon robot navigation without requiring 3D maps or pre-trained controllers. Our approach integrates global topological path planning with local metric trajectory control, allowing the robot to navigate towards object-level sub-goals while avoiding obstacles. We address key limitations of previous methods by continuously predicting local trajectory using monocular depth and traversability estimation, and incorporating an auto-switching mechanism that falls back to a baseline controller when necessary. The system operates using foundational models, ensuring open-set applicability without the need for domain-specific fine-tuning. We demonstrate the effectiveness of our method in both simulated environments and real-world tests, highlighting its robustness and deployability. Our approach outperforms existing state-of-the-art methods, offering a more adaptable and effective solution for visual navigation in open-set environments. The source code is made publicly available: \url{https://github.com/podgorki/TANGO}.

\end{abstract}

\section{INTRODUCTION}
Visual navigation is a fundamental challenge in robotics, with significant implications for autonomous agents operating in real-world environments. Traditional approaches often rely on constructing precise, globally consistent geometric 3D maps~\cite{mur2015orb, salas2013slam++, conceptgraphs}, which can be computationally intensive and difficult to generalize across diverse settings. Alternatively, methods designed for navigating in previously unseen environments~\cite{chaplot2020neural, hong2023learning} may not effectively leverage prior knowledge, limiting their efficiency and adaptability.

Inspired by human navigation abilities -- where we can traverse environments by reasoning over previously observed images or objects without detailed 3D maps -- visual topological navigation has emerged as a promising alternative~\cite{SPTM2018semi, shah2023gnm, RoboHop}. Recent research has predominantly focused on \textit{image-level} topological maps~\cite{SPTM2018semi, shah2023gnm}, which, while straightforward, have limited representational capacity. They often lack semantic richness and are sensitive to viewpoint changes, hindering their applicability in dynamic and diverse environments.

In contrast, \textit{object-level} topological maps~\cite{RoboHop} offer several advantages, including direct open-set natural language querying, semantic interpretability~\cite{keetha2021hierarchical}, and viewpoint-invariant visual recognition~\cite{garg24revisitanything}. These attributes are crucial for enabling open-world navigation that can be seamlessly deployed across different environments, tasks, and robotic platforms. However, integrating object-level topological information into navigation pipelines presents challenges, particularly in bridging global planning with local motion control while ensuring obstacle avoidance and traversability.

\begin{figure}
    \centering
    \includegraphics[width=1\linewidth, height=0.15\textheight]{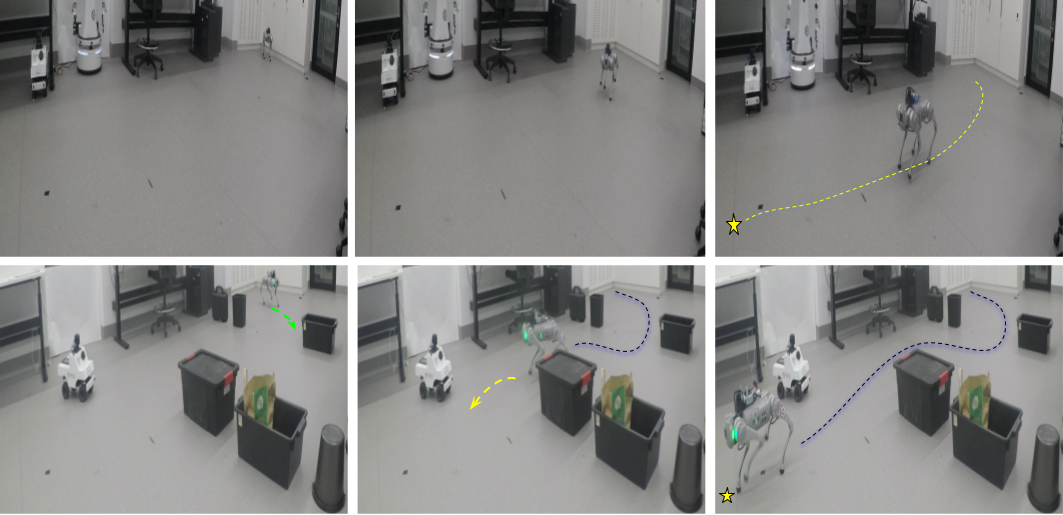}
    \caption{
    We present a topometric navigation pipeline that uniquely bridges \textit{topological} global path planner and \textit{metric} local trajectory planning, without needing 3D maps or learnt controllers. This enables our method to effective avoid obstacles (bottom row) even when no such objects were present in the mapping (teach) run.
    }
    \label{fig:teaser}
\end{figure}

In this work, we present a novel RGB-only, object-level, \textit{topometric} navigation pipeline for zero-shot robot control, in contrast with recent learnt controllers~\cite{cai2024bridging, SPTM2018semi, shah2023gnm}. Specifically, we propose a unique integration of global path planning and local motion planning, where a robot \textit{metrically} plans its motion to move towards \textit{topologically} planned object-level sub-goals. The latter is achieved through a recent work, RoboHop~\cite{RoboHop}, where its global path planner generates object-level sub-goal cost mask for robot's current observation (see Figure~\ref{fig:robohop+}). While this sub-goal mask can guide a robot for \textit{where to head}, it does not account for traversability or obstacle avoidance due to its purely topological nature. We address this limitation through our proposed topometric controller, where we explicitly predict traversable image segments, project them in Bird's Eye View (BEV) space using monocular metric depth, plan a trajectory to the farthest least-cost sub-goal, and continue this process until the long-horizon goal is reached.

The contributions of this paper are as follows:
\begin{itemize}
    \item a novel topometric controller that uniquely bridges \textit{topological} global path planner and \textit{metric} local trajectory planning to enable long-horizon object-goal navigation without needing 3D maps or learnt controllers;
    \item an RGB-only method to \textit{continuously} predict \textit{local trajectory} using single-view depth and traversability;
    \item an auto-switch-control approach that switches from our proposed controller to a fallback controller by detecting the absence of visible traversable regions; and
    \item a real-world demonstration (5 Hz) of our modular navigation pipeline built on top of `foundation models' such as Fast Segment Anything~\cite{zhao2023fast, kirillov2023segment}, Depth Anything~\cite{depthanything}, and CLIP~\cite{radford2021learning}, which \textit{explicitly} maintains an open-set applicability.
\end{itemize}

\section{RELATED WORKS}
\subsection{Topological sub-goals for Navigation}
A vast majority of vision-based navigation methods rely on 3D maps~\cite{huang23vlmaps, chaplot2020neural, conceptgraphs, salas2013slam++, rosinol2021kimera, georgakis2021learning}, where significant progress has been made both for `unseen'~\cite{zhang20233d, zhao2023zero, dorbala2023can, chaplot2020object} and prior map-based `seen'~\cite{weiss2011monocular, huang23vlmaps, TSGM} environments. In contrast, a range of methods exist that directly use visual topological sub-goals for long-horizon navigation, without requiring a 3D map. Inspired by human-like navigation capability, SPTM~\cite{SPTM2018semi} demonstrated a learning-based navigation controller using an image sequence as a map. Recent works in this direction have innovated in real-world training and deployment~\cite{shah2022viking}, use of language~\cite{shah2022lmnav}, adaptation to different embodiments~\cite{shah2023gnm}, and jointly learning to explore~\cite{sridhar2024nomad}. These methods predominantly use an image as a sub-goal, which has to have been captured from a different robot pose to obtain a control signal from the image pair. This can either be achieved through learning-based approaches~\cite{SPTM2018semi,shah2022lmnav,huang23vlmaps,li2020learning,meng2020scaling, ehsani2024spoc} or visual servoing~\cite{feng2021trajectory,bista2016appearance,mezouar2002path,hutchinson1996tutorial,cherubini2011visual,ahmadi2020visual,remazeilles20063d,diosi2011experimental,blanc2005indoor}. Instead of using image sub-goals, recent methods such as PixNav~\cite{cai2024bridging} and RoboHop~\cite{RoboHop} proposed to use respectively a pixel and an object as a sub-goal -- visible in the robot's current observation. While PixNav learns a controller in simulation for this purpose, RoboHop uses a zero-shot `segment servoing' approach to reach object sub-goals. In this work, we follow RoboHop's open-set navigation pipeline to obtain object sub-goals using its global path planner based on object-level topological graph, and propose to combine this with a traversability-aware trajectory planner to achieve a more performant navigation system.

\subsection{Open-set, Zero-shot, Large Models-enhanced Navigation}
There has been significant progress in learning-based navigation using both reinforcement~\cite{chen2018learning, wahid2021learning, bruce_sünderhauf_deepmind_london_deepmind_milford} and imitation~\cite{lee2021generalizable, pirlnav}, across different application areas including autonomous driving in structured environments~\cite{zhang2022rethinking, hu2023planning}, off-road outdoor settings~\cite{shah2022viking, jung2024v}, and aerial vehicles~\cite{loianno2020special, krishnan2021air, bonatti2020learning, weiss2011monocular}. With recent advances in large-scale general (foundation) models for perception, researchers have now focused on leveraging such models for their open-set characteristics and zero-shot applicability. Examples include ZSON~\cite{majumdar2022zson}, COW~\cite{gadre2023cows}, GOAT~\cite{chang2023goat, khanna2024goat}, and VL-Maps~\cite{huang23vlmaps}, which rely on joint vision-language embedding space of CLIP~\cite{radford2021learning} for open-vocabulary navigation. Although enabling open-set goal description in natural language, most of these methods rely on learning-based techniques for robot control.
In the same vein of using foundation models, Large Language Models (LLMs) have been explored for zero- or few-shot navigation, e.g., NavGPT~\cite{zhou2024navgpt, zhou2024navgpt2}, MapGPT~\cite{chen2024mapgpt}, and VisionGPT~\cite{wang2024visiongpt}. More recently, multimodal LLMs have also been leveraged for navigation using videos, e.g., NaVid~\cite{zhang2024navid} and MobilityVLA~\cite{chiang2024mobility}, and visual annotations, e.g., PIVOT~\cite{nasiriany2024pivot} and CoNVOI~\cite{sathyamoorthy2024convoi}. While these methods aim to directly control the robot actions, they are limited in terms of their grounding in the robot's map~\cite{zhou2024navgpt}, 3D spatial understanding~\cite{nasiriany2024pivot}, and long inference times~\cite{chiang2024mobility, zhang2024navid}. Distinct from the aforementioned techniques, we build a \textit{modular} open-set navigation pipeline to generate zero-shot control signal by using Segment Anything Model (SAM)~\cite{kirillov2023segment} for object-level topological mapping and planning; CLIP~\cite{radford2021learning} combined with SAM for traversability estimation; and Depth Anything~\cite{depthanything} for monocular depth estimation of traversable segments for local motion planning in BEV space.

\subsection{Teach \& Repeat and Experiential Navigation}
A significant subset of navigation literature deals with visual teach-and-repeat (VT\&R) task~\cite{furgale2010visual,vsegvic2009mapping,zhang2009robust,dall2021fast,mattamala2022efficient,krajnik2018navigation,halodova2019predictive,do2019high,krajnik2017image}. These methods typically do not require 3D maps for navigation, as they implicitly leverage the inherent assumption of the `narrower' task of repeating the teach run by simply using image-based visual servoing. In order to avoid collisions with obstacles, such navigation pipelines explicitly estimate free space locally~\cite{bista2021image}, or estimate traversability from prior map information~\cite{mattamala2022efficient, berczi2017looking}, often using depth sensors~\cite{bista2021image, mattamala2022efficient}. A more generalized version of such navigation can be referred to as experiential learning of robot navigation~\cite{levine2023learning}. By learning from real-world data at large-scale, e.g., ViNT -- a foundation model for navigation~\cite{shah2023vint}, such control policies can exhibit general understanding of traversability, reachability, and exploration objectives. Although more capable than teach-and-repeat, the end-to-end learning paradigm of such methods limits their interpretability, and induces control-related data bias that can limit their broader applicability.  In contrast to learning `foundational control', we aim to leverage `vision foundations for navigation', which presents several benefits: open-set, object-level queryable map; interpretable object-level global plan; and explicit traversability-aware local motion planning. This also enables a novel capability: reaching \textit{seen but unvisited} object goals, which steps beyond simple teach-and-repeat task while not requiring any learning. 

\begin{figure*}
    \centering
    \includegraphics[width=0.95\textwidth, height=0.25\textheight]{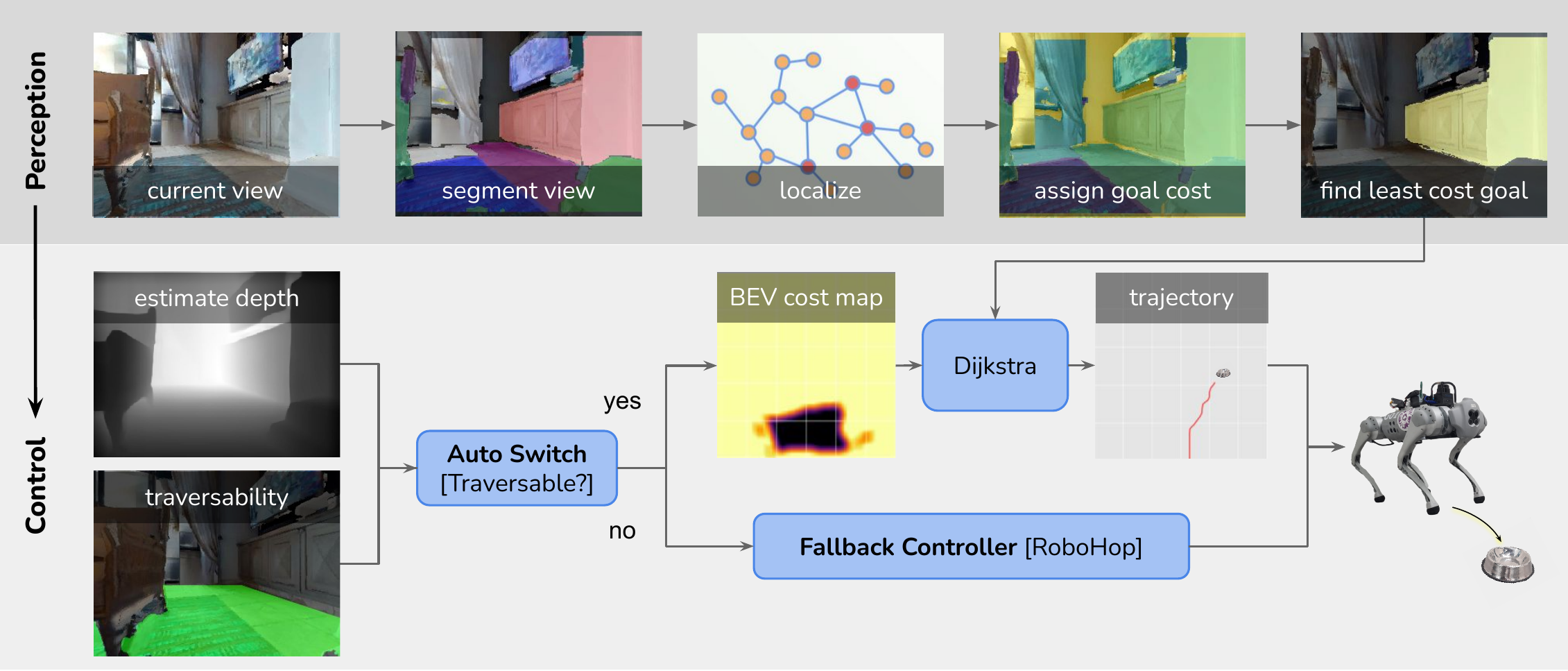}
    \caption{\ourM{}'s Navigation Pipeline.
    \textbf{Perception}: The robot's current view is segmented using a foundational segmentation model (SAM), the segments are localised within an object-level topological map using local feature matching (LightGlue). Each segment is assigned a cost based on its topological proximity to the final goal segment, the segment closest to the final goal is selected to drive the controller.
    \textbf{Control}: A BEV traversability map is computed by combining state-of-the-art depth estimation with open-set text query capabilities (CLIP) to identify traversable surfaces such as `floor' or `ground'. This depth and semantic information is integrated to generate a BEV cost map (yellow high cost, black low cost). Dijkstra's algorithm is applied to compute the shortest path to the sub-goal segment, providing a trajectory that avoids obstacles and generates yaw control signals for robot navigation. This perception-action loop is repeated continuously until the robot reaches the final goal object.}
    \label{fig:robohop+}
\end{figure*}

\section{APPROACH}
Our proposed method aims to effectively integrate the robot's understanding of topologically-connected object sub-goals with an ability to reach that sub-goal through traversability-aware instantaneous trajectory planning. As the topological global planner continuously updates the sub-goal masks, the metric local planner enables movement through the traversable paths to continuously reach the updated sub-goals until the final goal is reached. We present our topometric controller in the following subsections, where we first provide background details for the topological mapping, localization and planning pipeline based on RoboHop~\cite{RoboHop}, and then describe the novel integration of our local metric motion planner. 

\subsection{Topological Object-based Mapping and Planning}
\label{sec:topo_mapping}

We define the topological map of the environment as a graph $\mathcal{G} = (\mathcal{N},\mathcal{E})$, where $\mathcal{N}$ and $\mathcal{E}$ denote the set of nodes and edges, respectively. Each node $n_i$ in $\mathcal{G}$ corresponds to an image segment $\textbf{M}_i$, which represents a meaningful object. 
An edge $e_{ij} \in \mathcal{E}$ connects image segments and is defined as either: a) \textit{intra-image edges}, which connect centroids $\textbf{M}_i$ and $\textbf{M}_j$ within the same image $I^t$ using Delaunay Triangulation, or  b) \textit{inter-image edges} which connect corresponding segments matched across different images through data association.

\subsubsection*{Mapping} The segmentation masks are extracted from an image sequence $\{I^t\}$ using a foundational model, such as SAM~\cite{kirillov2023segment}. The zero-shot capabilities of these recent foundational models are particularly valuable, as they enable us to construct a topological representation that is not restricted to a closed-world assumption of predefined objects. Moreover, these models inherently support integration with richer descriptors and language models, allowing for more expressive scene understanding. For node/segment tracking during the mapping process, we utilise local feature matching, which was observed to perform better than DINOv2-based matching in the original RoboHop~\cite{RoboHop}. Specifically, we extract SuperPoint~\cite{detone2018superpoint} features and match them using LightGlue~\cite{lindenberger2023lightglue} to identify pixel-level correspondences between an image pair. These matches are converted to segment-level correspondences based on the membership of the pixels in their respective segmentation masks.

\subsubsection*{Localisation} At every step, the robot localises itself within a temporal window of map images centered around its previous estimation of localised reference image index. Given the candidate map images, the robot's current image is matched pairwise using the same local feature matching process described in the mapping section above. This provides segment-level correspondences between the current image and the map images. Using these correspondences, we obtain sub-goal costs for each of the query segments using the global planner, as described next.

\subsubsection*{Global Planning} 
By leveraging the connectivity between segments in the map, we compute path lengths between the localized reference map segments and the goal segment. To facilitate this, we assign edge weights between the source and destination nodes: specifically, inter-image edges (connecting segments from different images) are assigned a weight of 0 (being the same object instance), while intra-image edges (connecting segments within the same image) are given a weight of 1. We then compute a weighted shortest path to the target node in the map using Dijkstra's algorithm, starting from every localized query segment. This yields a sub-goal cost mask for the robot's current observation (see Figure~\ref{fig:robohop+}), highlighting the desirable objects that the robot should approach to reach its long-horizon goal.

\subsection{Metric Control To Reach Object Sub-Goals}
\label{sec:metric_control}

Given the object-level sub-goals planned topologically by the global planner, \ourM{} generates a local metric motion plan to navigate toward these sub-goals. The transition from topological sub-goals to metric sub-goals is accomplished by computing a BEV traversability map. Using state-of-the-art models, our method combines single-view depth estimation with open-set text query capabilities, enabling the refinement of traversable regions based on object semantics.

\subsubsection*{Metric BEV Traversability} At each timestep the robot's RGB image is converted to binary segment masks using a foundational model such as SAM \cite{kirillov2023segment}. Each segment is assessed for traversability by utilising CLIP~\cite{radford2021learningtransferablevisualmodels} text queries, filtering out segments based on their ``semantics'', such as floor, ground, or rug. Segments within the segment map are set to 1 if assessed as traversable and 0 if non-traversable forming a binary traversability mask. This open-set queryable filter enables fine-grained selection of traversable regions, adaptable to different real-world scenarios.
For each sub-goal node segment $n_i$, we select the lowest-cost image segment $\textbf{M}_i$ as the representative sub-goal. Once the traversable segment masks and the sub-goal segment are identified, we apply monocular depth estimation via Depth-Anything~\cite{depthanything} to project the traversable segments and sub-goal points into 3D space, resulting in the final metric BEV traversability map. The final sub-goal-point is then selected as the farthest projected point contained in the sub-goal-segment.

\subsubsection*{Trajectory and Motion Planning} For each input RGB image frame, the metric BEV traversability map is calculated and converted to a cost map for planning. The cost map is formed by applying a distance transform from the traversability masks edges, which is then smoothed with a box filter. Within this cost map, the shortest path to the local 3D sub-goal is determined using Dijkstra's algorithm, generating a series of traversable waypoints along the trajectory to the sub-goal. These waypoints are then used to generate control signals, controlling the robot's yaw angle with a proportional line following controller and holding the linear velocity fixed to effectively navigate toward the sub-goal.

\subsubsection*{Auto Switch Control} In situations where metric traversability prediction is unreliable or unavailable, such as when the robot is too close to a wall or obstructed by an obstacle, the local controller automatically switches to RoboHop's~\cite{RoboHop} fully topological ``segment servoing'' approach. In the absence of reliable traversable regions, this method converts the horizontal pixel offset of each sub-goal mask into yaw velocity ($\theta$) using Equation~\ref{eq:robohop}, ensuring the robot can still navigate effectively in these challenging scenarios.
\begin{equation}
    \theta = \frac{G}{W} \sum_i w_i(u_i - c_i)
    \label{eq:robohop}; \quad w_i = \frac{e^{\tau l_i}}{\sum_i{e^{\tau l_i}}}
\end{equation}
where $c_i$ represents the image centre, $u_i$ is the centroid of the segment $\textbf{M}_i$, $l_i$ is the path length (min-max normalized across localized query segments), $\tau$ is the temperature parameter (set to $5$), $w_i$ is the softmax weight per query segment, $W$ is the image width, and $G$ is the controller gain (set to $0.4$).

\section{EXPERIMENTS}
\subsection{Dataset and Evaluation}
We use Habitat-Matterport 3D Dataset (HM3D)~\cite{Ramakrishnan2021HabitatMatterport3D} to evaluate our proposed method. Specifically, we use the validation set of the InstanceImageNav (IIN) challenge~\cite{krantz2022instance} that comprises $36$ unique environments. We sample $3$ episodes (with unique object goals) per environment to benchmark across $108$ episodes. For each episode, we use the simulator's path finding method to obtain a mapping (teach) traverse, which is used to construct the object-level topological graph and is available during inference to all the methods for generating sub-goal costs.

We evaluate a controller's ability to navigate to an object goal in a given episode. We report the average success rate, where an episode is deemed successful if the robot is within $1m$~\cite{habitatchallenge2023} of the goal position, taking maximum $500$ steps. The evaluation is repeated based on the starting point of the robot by varying the geodesic length of trajectories. While PixNav~\cite{cai2024bridging} only uses two short variations of episode lengths, we further include the full length of the episodes as a more challenging setting. We refer to these as `easy', `hard' and `full', with their starting distance from the goal respectively as $1$-$3m$, $3$-$5m$, and $8$-$10m$. 
We provide all implementation details in the supplementary video, along with real-world demonstrations.
\subsection{Baselines}
We use the following baselines to assess the effectiveness of our proposed method.
\subsubsection{Ground Truth Goal Masks} We consider two key variants of our navigation pipeline where we use ground truth information for perception and planning to generate goal masks corresponding to robot's image observation. i) \texttt{GT-Metric}: we use simulator's semantic instance masks, depth and navigation mesh to find shortest (geodesic) paths from each object instance to the goal object. This provides an accurate metric estimate of object sub-goal costs in robot's current view, thus being the ideal goal mask input for the controller. ii) \texttt{GT-Topological}: we use simulator's semantic instance masks to create an object-level topological map (as described in Section~\ref{sec:topo_mapping}), which assumes segmentation, matching/association, and localization to be solved. This object-level map is then used to compute global path lengths, thus the goal masks so obtained lack geometric understanding of object segments layout and only rely on intra-image and inter-image object connectivity. This setting aids in testing the role of planning as well as control while assuming perception to be solved.

\subsubsection{Robohop}
We use the original RoboHop controller as described in Eq.~\ref{eq:robohop}, where the goal mask information is used in the form of pixel centers of the object segments weighted by their path lengths.

\subsubsection{Pixel-guided Navigation (PixNav)}
PixNav is a transformer-based imitation learning local navigation method~\cite{cai2024bridging} that uses a patch of goal pixels that correspond to the final or intermediate navigation goal points. The goal patch is initially fed into the model as a mask with the corresponding RGB image and then executes an action from the discrete action space:  {Stop, MoveAhead, TurnLeft, TurnRight, LookUp, LookDown}. At each subsequent step the current RGB image and a collision signal are used with the history of the previous images and the initial goal mask to predict the next action. 

PixNav is a discrete controller with a move-able camera whereas RoboHop and \ourM{} are continuous with a fixed camera. 
Accounting for these differences and noting the intended design of PixNav, evaluations for the PixNav controller were set to initialize a viewable intermediate goal given by the topological global planner, where the goal was updated when the method outputted `Done' or when its memory buffer was full. 

\section{RESULTS}

\subsection{Benchmark comparison}
Table~\ref{tab:res_main} presents a comparison of our proposed method against baseline techniques for varying lengths of trajectories. We also include ablative results where we swap the perception and planning modules with their ground truth counterparts. All compared methods are provided the same goal information to observe the control performance differences in isolation. For GT-Metric and GT-Topological settings, \ourM{} uses simulator's depth and traversability estimation, we ablate this in Section~\ref{subsec:ablate_gt} for further insights. It can be observed that our proposed method \ourM{} outperforms both the baselines -- learning-based controller PixNav and traversability-unaware zero-shot controller RoboHop -- by a significant margin in most cases. The trend remains consistent across easy, hard, and full length trajectories. Furthermore, as the extent of ground truth information reduces from \texttt{GT-Metric} through \texttt{GT-Topological} to \texttt{No-GT}, a general trend of gradual performance decline is observed. For the \texttt{GT-Metric} setting, `perfect' sub-goal masks lead to the highest navigation success rate for all methods, as expected. For the \texttt{GT-Topological} setting, performance drops become large for the hard and full-length trajectories. These results highlight the impact of topologically computed global path lengths in contrast with \texttt{GT-Metric} which assumes access to a full geometric 3D map/simulator. In the \texttt{No-GT} setting, although the absolute performance of all methods is low, the comparison to its \texttt{GT} counterparts shows that the performance drop is attributed more to perception than topological planning and control.  

\begin{table}
\centering
\caption{Navigation success rate across varying trajectory lengths.}
\begin{tabular}{cccc} 
\toprule
 Controller& Easy [1-3m] & Hard [3-5m]& Full [8-10m]\\
 \midrule
  &\multicolumn{3}{c}{GT-Metric}\\
  \cmidrule(lr{0.75em}){2-4}
     RoboHop~\cite{RoboHop}& 
 93.14 & 78.43&   42.16\\
 PixNav~\cite{cai2024bridging}& 65.69 & 44.12 & 15.69 \\  %
 \ourM{} (ours)& \textbf{94.12} & \textbf{90.20}&   \textbf{48.04}\\
 \midrule
 &\multicolumn{3}{c}{GT-Topological}\\
  \cmidrule(lr{0.75em}){2-4}
 RoboHop~\cite{RoboHop}& \textbf{78.43}& 58.82 & 25.49\\ %
 PixNav~\cite{cai2024bridging}& 60.78 & 44.12 & 15.69 \\  %
 \ourM{} (ours) & 74.51& \textbf{65.69}& \textbf{30.39}\\ %
 \midrule
 &\multicolumn{3}{c}{No-GT}\\ 
   \cmidrule(lr{0.75em}){2-4}
  RoboHop~\cite{RoboHop}& 43.56 & 34.56 & 13.73 \\ %
 PixNav~\cite{cai2024bridging} & 51.96 & 39.22 &  14.0 \\ %
    \ourM{} (ours) & \textbf{61.76} & \textbf{43.14} & \textbf{21.57} \\ %
 \bottomrule
\end{tabular}
    \label{tab:res_main}
\end{table}

\subsection{Ablation Studies}
\subsubsection{Control with and without GT}
\label{subsec:ablate_gt}
In Table~\ref{tab:res_ablation}, we compare different versions of our proposed controller by ablating ground truth (simulator) components of perception (segmentation and association) and control (depth estimation and traversability) with their respective prediction methods. It can be observed that our full pipeline (last row) only suffers $5\%$ performance drop in comparison to the simulated control components (depth and traversability), whereas perception (and localization) prediction leads to a $18\%$ drop (first to second row). These comparisons emphasize the need for improved segmentation and matching methods more than monocular depth estimation for the downstream task of navigation.

\begin{table}
\centering
\caption{Ablation navigation success rate for the \ourM{} controller across 'hard' 3-5m trajectories.}
\begin{tabular}{ccc>{\centering\arraybackslash}p{1.3cm}}
\toprule
  Perception & Control & \multirow{2}{*}{Success Rate}\\
 \cmidrule(lr{0.75em}){1-1}
 \cmidrule(lr{0.75em}){2-2}
  Segment + Matcher & Depth + Trav &  \\ 
 \midrule
    Sim& Sim & 65.69\\ %
    FastSam + LGlue& Sim & 47.95\\ %
  FastSam + LGlue& DepthAnything + FastSam & 43.14 \\ %

\bottomrule
\end{tabular}
    \label{tab:res_ablation}
\end{table}

\subsubsection{Auto Switch Control}
The proposed local controller creates a relative bird's-eye-view based on the traversability of the scene at each step. This enables traversal around objects blocking robot's path. However, in tight spaces of a house, there exists situations where the controller is unable to perceive traversable segments, or the traversability estimation fails. In these cases, object segments can still be observed and matched to obtain a valid goal to control the robot's yaw. Thus, the proposed local motion planner switches to the RoboHop controller as fallback, which rotates the robot towards the goal until the traversable segments are visible again for the proposed controller to take over. Table~\ref{tab:res_fallback} presents a comparison of our local motion planner with and without the proposed auto switching. The controller variations were evaluated in the \texttt{GT-metric} `hard' setting for a maximum of 250 simulation steps, differing from the other evaluations which use 500 steps. It can be seen that in times of unknown traversability, having the ability to fall back to ``segment servoing'' enables continual progress towards the goal, thus improving the success rate.

\begin{table}
\centering
\caption{Auto Switch Control improves \ourM's success rate.}
\begin{tabular}{cccc} 

\toprule
Control Type & No Switch & Auto Switch & Improvement \\
\midrule
Hard [3-5m] & 62.14 & 73.78 & \textbf{11.64} \\
\bottomrule
\end{tabular}
    \label{tab:res_fallback}
\end{table}

\begin{table}
\centering
\caption{Reaching Seen-but-Unvisited Goals.}
\begin{tabular}{ccc} 
\toprule
Goals Type & Hard [3-5m] & Full [8-10m] \\
\midrule
Teach Goals & 43.14 & 21.57 \\
Alt Goals & 50.54 & 25.84 \\
\bottomrule
\end{tabular}
    \label{tab:res_seenButUnvisited}
\end{table}

\begin{figure}
    \centering
    \includegraphics[scale=0.3, trim={2.5cm 2.5cm 2.5cm 2.5cm}, clip]{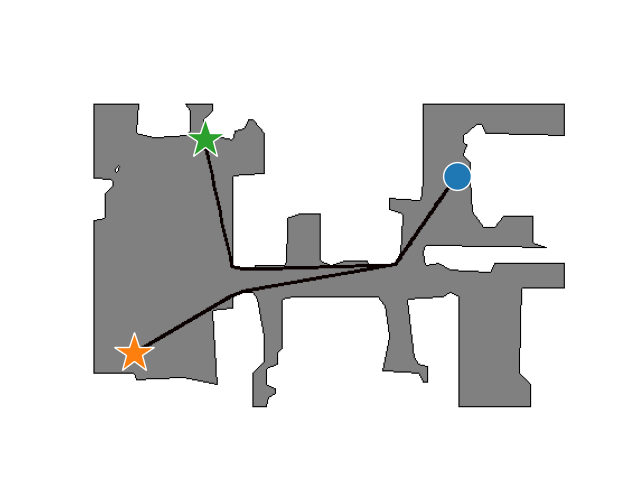}
    \caption{Seen-but-Unvisited Goals: An example episode with agent's starting position marked in blue, original goal in green and the new seen-but-unvisited goal in orange. The robot has no `prior experience' of reaching the new (orange) goal, as the map run is traversed from blue to green.}
    \label{fig:alt_goal}
\end{figure}

\subsection{Reaching `seen but unvisited' Goals}
Our proposed method uses a \textit{topological} prior map based on a single trajectory. The objects on the way can thus be assumed to be reachable. However, several more objects (in other rooms/places) are observed throughout the mapping run, which the robot may not have `prior experience' of reaching. These \textit{seen-but-unvisited} objects can be selected as a new alternate goal -- reaching these goals using only an image-level connectivity may not be possible, whereas object-level connectivity enables identifying and reaching these goals through a traversability-aware local metric controller. Thus, stepping beyond the typical teach-and-repeat paradigm, we evaluate our method's navigation success in reaching \textit{seen-but-unvisited} long-horizon object goals. For each of the episodes, we obtain such goals, referred to as \texttt{Alt Goals}, through a simple measure: an object instance (excluding the \texttt{wall} and \texttt{ceiling} class) is chosen from the last 30\% the episode's map (teach) run poses such that the sum of object's average depth from a given pose and its geodesic distance to the original goal is the highest across all possible instances. Thus, the \texttt{Alt Goals} get sampled from different rooms or in the same room but far-off from the original goal, as shown in Figure~\ref{fig:alt_goal}. We evaluate on this new task exactly as described for the vanilla task except that the original object goals are replaced by these new goals.

In Table~\ref{tab:res_seenButUnvisited}, it can be observed that \ourM's success rate for reaching seen-but-unvisited \texttt{Alt Goals} is comparable to that for \texttt{Teach Goals}, across both `hard' and `full' setting. While this clearly demonstrates the capability of our proposed pipeline beyond simple teach-and-repeat, the performance comparison highlights that low performance is not attributed to the difference in the task but to the limitations of perception and planning, as established in Table~\ref{tab:res_main}. Overall, these results emphasize the role of \textit{object-level} (thus, traversability-aware) topological maps for navigation as opposed to their image-level counterparts, where the latter's lack of an explicit reasoning of objects and traversability limits its goal-reaching to only the poses the map images were captured from.

\subsection{Qualitative Analysis}
\begin{figure}
    \centering
    \includegraphics[width=0.475\textwidth, height=0.15\textheight]{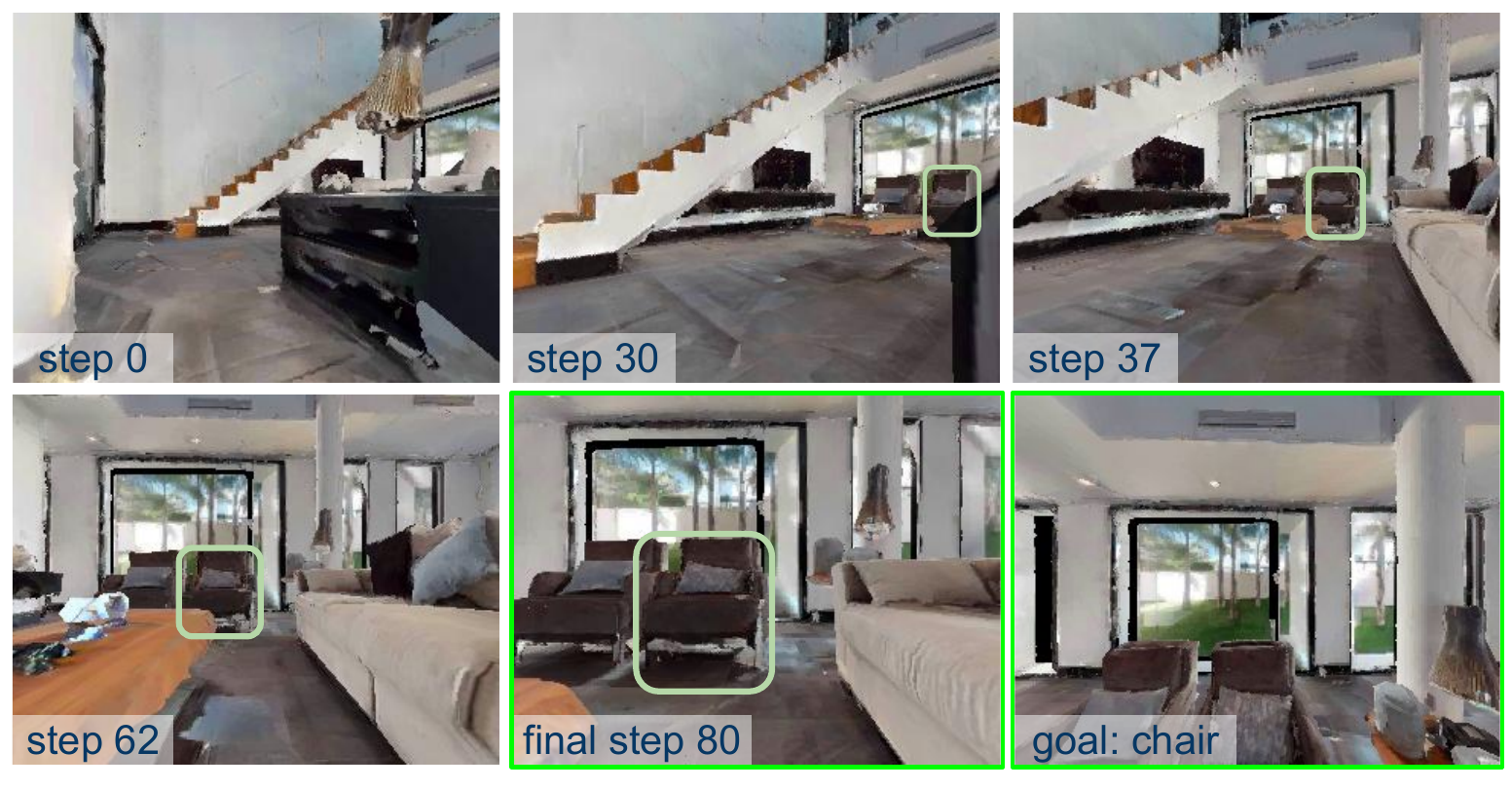}
    \caption{Successful sequence showing the controller navigating around the couch and between the table successfully arriving at the final goal chair (highlighted in green box)}
    \label{fig:win_000}
\end{figure}
\begin{figure}
    \centering
    \includegraphics[width=0.475\textwidth, height=0.15\textheight]{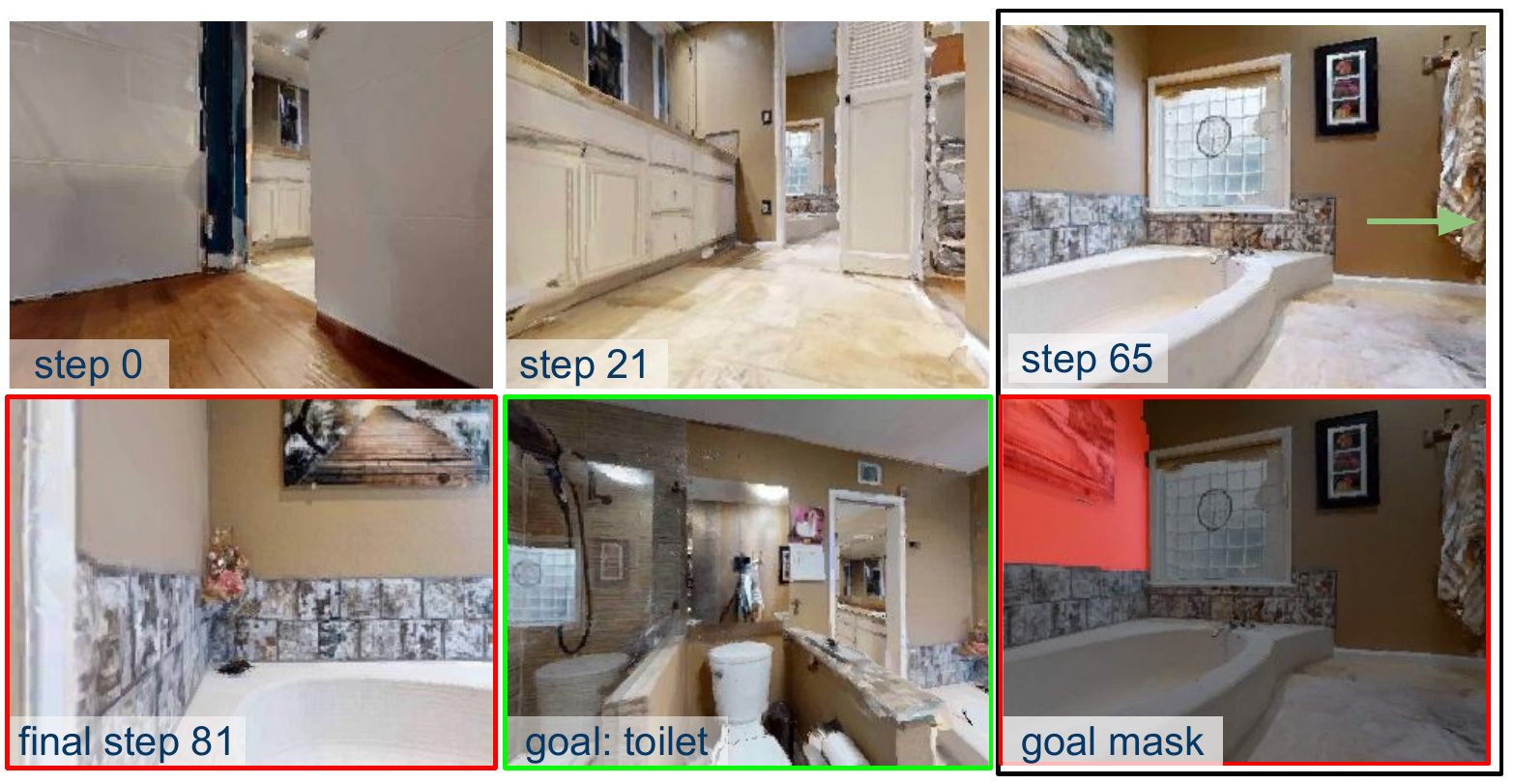}
    \caption{Unsuccessful sequence showing the controller navigating around into the bathroom and incorrectly turning towards the vase at step 65 rather than correctly steering towards the toilet - goal location is indicated with green arrow. Lower right side figure shows overlaid (red) goal segment which steers the controller to the left rather than the right.}
    \label{fig:fail_000}
\end{figure}

Under successful trajectories such as shown in Figure \ref{fig:win_000} the controller displays desirable behaviours such as, an ability to turn correctly, avoid objects and choose a path between objects that are close together such as a coffee table and a couch. On occasion misleading goals from the topological global planner will cause the robot to move in a less optimal direction which results in the robot moving around for an increased number of steps. However, a common failure mode of the local planner is shown in Figure \ref{fig:fail_000}. In this particular instance, the controller successfully traverses towards the goal from one room turning into another, following through this room until reaching the final bathroom. In the bathroom, the controller continues to plan and control correctly until an erroneous goal segment influences the controller to turn left rather than right shown in the lower right in white. This error mode highlights the importance of high quality matches between the current view segments and the topological map.

\section{CONCLUSIONS}
Topological visual goals based navigation using a single RGB camera is an appealing alternative to classical methods based on 6-DoF pose estimation and geometrically-precise 3D maps. This paper presents a novel topometric navigation controller that bridges object-level topological global planning with traversability-aware local metric motion planning using instantaneous monocular depth. This unique integration built on top of `vision foundation models' leads to a more readily-deployable navigation system, which performs significantly better than the previous methods including both a learnt and zero-shot controller. Consequently, we demonstrate an interesting a new navigation capability of reaching `seen-but-unvisited' object goals, which emphasizes the importance of a ground-up object-level navigation pipeline. Furthermore, we demonstrate real-world experiments showcasing obstacle avoidance under significant changes in the map (teach) run. 

There are a few limitations of our pipeline which lead to navigation failures: a) Perception: incorrect matching of segments from the current view to the reference segment map leads to incorrect sub-goals; b) Planning: pure-topology based edges in the map graph lack the ability to geometrically disambiguate the relevance of different sub-goals in the current image; and c) Traversability: text- and segmentation-based estimation, although convenient, is prone to errors which leads to the use of a fallback controller. However, as opposed to end-to-end learnt controllers, the modular nature of our proposed pipeline allows drop-in replacement of different components as more performant perception models become rapidly available in future.

\bibliographystyle{IEEEtran}
\bibliography{references}

\end{document}